%% file: Main.tex
\documentclass[sigconf]{acmart}

\usepackage{algorithm}
\usepackage{algorithmic}

\usepackage{newfloat}
\usepackage{listings}

\usepackage{amsmath}

\usepackage{amssymb}
\usepackage{booktabs} 
\usepackage{multirow}
\usepackage[most]{tcolorbox}
\usepackage{xcolor}
\usepackage{lipsum}
\usepackage{color}

\newcommand{\up}[1]{\textsubscript{\footnotesize \textcolor{red}{$\uparrow$#1}}}
\newcommand{\down}[1]{\textsubscript{\footnotesize \textcolor{blue}{$\downarrow$#1}}}

\setcopyright{acmlicensed}
\copyrightyear{2026}
\acmYear{2026}
\setcopyright{cc}
\setcctype{by}
\acmConference[KDD '26]{Proceedings of the 32nd ACM SIGKDD Conference on Knowledge Discovery and Data Mining V.2}{August 09--13, 2026}{Jeju Island, Republic of Korea}
\acmBooktitle{Proceedings of the 32nd ACM SIGKDD Conference on Knowledge Discovery and Data Mining V.2 (KDD '26), August 09--13, 2026, Jeju Island, Republic of Korea}
\acmDOI{10.1145/3770855.3817987}
\acmISBN{979-8-4007-2258-5/2026/08}

\begin{document}

\title{TuneAgent: Agentic Operating System Kernel Tuning with Reinforcement Learning}


\author{Hongyu Lin}
\authornote{Both authors contributed equally to this research.}
\email{hongyu2021@iscas.ac.cn}
\affiliation{
\institution{Institute of Software, Chinese Academy of Sciences}
\institution{University of Chinese Academy of Sciences}
\state{Beijing}
\country{China}
}

\author{Yuchen Li}
\email{liyuchen2021@iscas.ac.cn}
\authornotemark[1]
\affiliation{
\institution{Institute of Software, Chinese Academy of Sciences}
\institution{University of Chinese Academy of Sciences}
\state{Beijing}
\country{China}
}

\author{Haoran Luo}
\authornote{Corresponding authors.}
\email{haoran.luo@ieee.org}
\affiliation{
\institution{Nanyang Technological University}
\country{Singapore}
}

\author{Zhenghong Lin}
\email{hongzhenglin970323@gmail.com}
\affiliation{
\institution{Nanyang Technological University}
\country{Singapore}
}

\author{Libo Zhang}
\email{libo@iscas.ac.cn}
\affiliation{
\institution{Institute of Software, Chinese Academy of Sciences}
\institution{University of Chinese Academy of Sciences}
\state{Beijing}
\country{China}
}

\author{Mingjie Xing}
\authornotemark[2]
\email{mingjie@iscas.ac.cn}
\affiliation{
\institution{Institute of Software, Chinese Academy of Sciences}
\institution{University of Chinese Academy of Sciences}
\state{Beijing}
\country{China}
}

\author{Yanjun Wu}
\email{yanjun@iscas.ac.cn}
\affiliation{
\institution{Institute of Software, Chinese Academy of Sciences}
\institution{University of Chinese Academy of Sciences}
\state{Beijing}
\country{China}
}


\renewcommand{\shortauthors}{Hongyu Lin et al.}

\begin{abstract}
Linux kernel tuning is essential for optimizing operating system (OS) performance, yet remains challenging due to the complex kernel space, sparse performance feedback, and strong workload sensitivity. We present \textbf{TuneAgent}, an agentic Linux kernel tuning framework powered by rule-based reinforcement learning (RL). TuneAgent formulates the kernel space as a constrained RL environment, enabling large language models (LLMs) to autonomously explore the kernel while enforcing valid and precise configuration modifications. To address sparse performance feedback, we design structured reward functions that jointly promote reasoning standardization, configuration correctness, and performance awareness. Furthermore, we propose a two-phase training strategy that first ensures format and semantic correctness and then transitions to performance-driven exploration, accelerating convergence and reducing overhead. Experimental results show that TuneAgent consistently outperforms existing baselines, achieving up to 5.6\% relative overall performance improvement while maintaining high configuration validity. We further demonstrate its robustness across multiple real-world applications, highlighting its practicality and adaptability in diverse deployment environments.
\end{abstract}


\ccsdesc[500]{Software and its engineering~Operating systems}
\ccsdesc[300]{Software and its engineering~Software creation and management}
\ccsdesc[500]{Computing methodologies~Reinforcement learning}
\ccsdesc[500]{Computing methodologies~Intelligent agents}
\ccsdesc[300]{Computing methodologies~Dynamic programming for Markov decision processes}

\keywords{Operating System Kernel Tuning, Large Language Model-based Agent, Reinforcement Learning}


\maketitle

\input{./Section/Introduction}
\input{./Section/RelatedWork}

\input{./Section/Dataset}
\input{./Section/Methdology}
\input{./Section/Experiment}

\input{./Section/Conclusion}

\begin{acks}
This work was supported by the Key Research Program of Frontier Sciences, CAS, Grant No. ZDBS-LY-JSC038.
\end{acks}

\bibliographystyle{ACM-Reference-Format}
\bibliography{kdd2026}
\input{./Section/Appendix}

\end{document}

%% file: Section/Introduction.tex
\section{Introduction}
\begin{figure}[t]
\centering
\includegraphics[width=70mm]{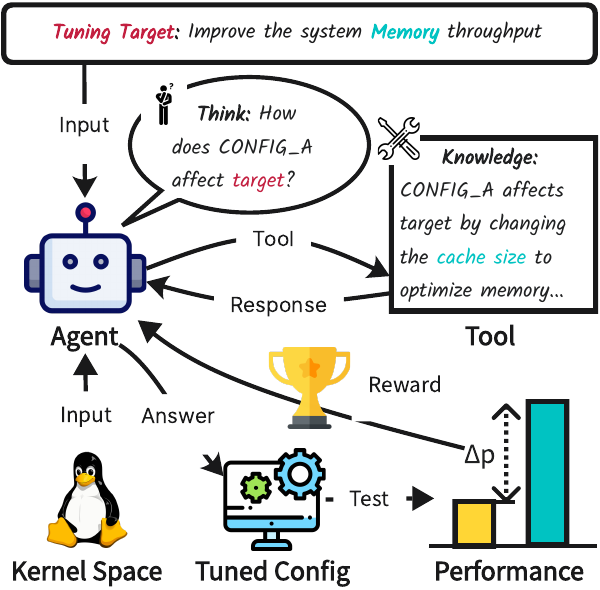}
\caption{An illustration of Linux kernel tuning using a reinforcement learning-trained large language model.}
\Description{An illustration of Linux kernel tuning using a reinforcement learning-trained large language model.}
\label{fig:task}
\vskip -0.1in
\end{figure}

The Linux kernel, as the core of modern operating systems (OSs), plays a pivotal role in determining overall system performance~\cite{linux}. As workloads become increasingly diverse and resource-intensive, \textbf{kernel tuning}—which involves systematically adjusting kernel configurations to optimize performance for specific workloads—has become an essential yet challenging task (Figure~\ref{fig:task}). Effective kernel tuning requires not only deep expertise in kernel internals but also a precise understanding of workload characteristics, making the process highly difficult to automate in practice.

\begin{figure*}[htbp]
\centering
\includegraphics[width=\textwidth]{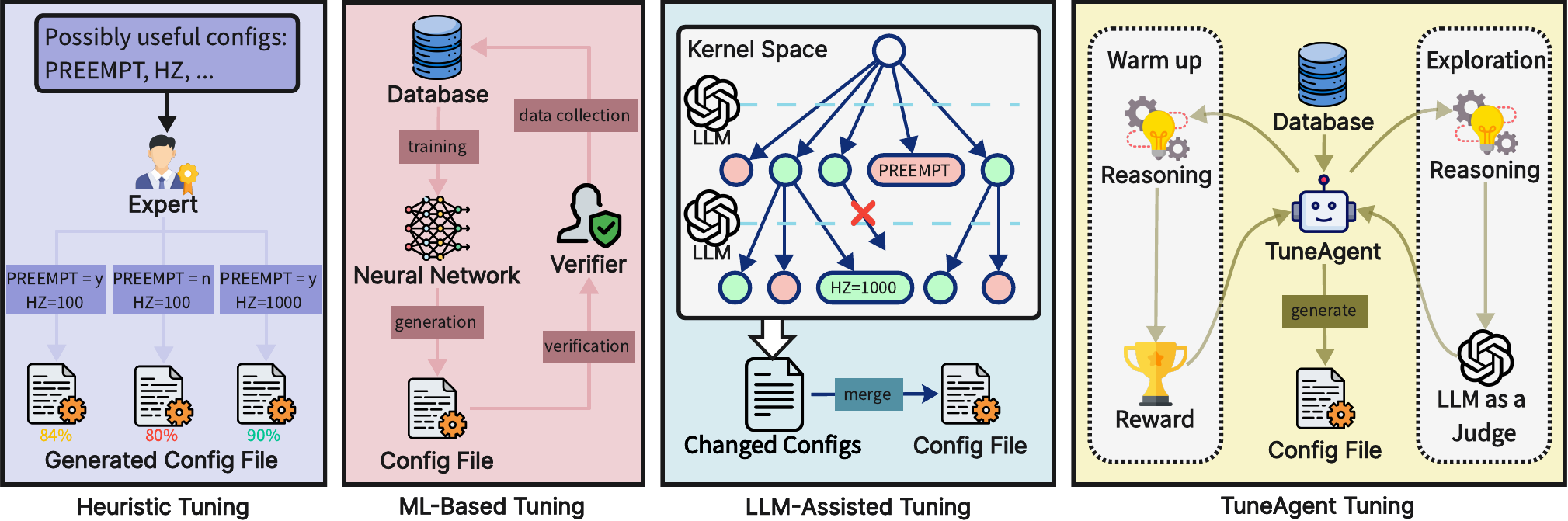}
\caption{Comparison of different kernel tuning methods, including heuristic-based tuning through variable control, ML-based tuning using neural networks, LLM-assisted tuning with tree-based search, and our proposed TuneAgent method, which incorporates RL and agent-based optimization.}
\Description{Comparison of different kernel tuning methods.}
\label{fig:comparison}
\end{figure*}

Despite extensive prior efforts, kernel tuning remains labor-intensive and error-prone. As illustrated in Figure~\ref{fig:comparison}, \textbf{heuristic tuning} relies heavily on expert knowledge to manually adjust kernel configurations~\cite{braswell2008linux, kernel1}. While effective in certain scenarios, these approaches are time-consuming and overhead-intensive. \textbf{Machine learning (ML)-based} methods attempt to automate tuning through data-driven optimization~\cite{kernel4, kernel3}, but often suffer from high data requirements and limited scalability. More recently, \textbf{large language model (LLM)-assisted tuning} frameworks leverage LLMs to reason about workload demands and propose kernel configurations~\cite{chenautoos}. However, such methods still struggle to efficiently navigate the constrained kernel space and ensure the validity and effectiveness of the generated configurations.

More fundamentally, Linux kernel tuning poses several intrinsic challenges that stem from its OS-level nature. \textbf{(1) Highly constrained and structured configuration space.} The kernel exposes over 18,000 configuration options governed by complex dependency, hierarchy, and mutual-exclusion constraints. Invalid combinations can lead to compilation failures or system crashes, making unconstrained exploration infeasible. \textbf{(2) Sparse and indirect performance feedback.} Unlike lightweight parameter tuning, kernel tuning requires rebuilding, deploying, and benchmarking the system to observe performance changes. Performance signals are sparse, delayed, and workload-dependent, significantly complicating the design of effective feedback mechanisms. \textbf{(3) Strong workload sensitivity and limited transferability.} Kernel configurations that benefit one workload may degrade performance for another, necessitating adaptive strategies that generalize across diverse workloads while minimizing retraining costs.

To address these challenges, we propose \textbf{TuneAgent}, an agentic framework for Linux kernel tuning powered by \emph{rule-based reinforcement learning (RL)}, where each key design component explicitly targets a core challenge of kernel tuning. 
\textbf{Kernel tuning as an RL problem:} TuneAgent abstracts the constrained kernel configuration space into an interactive RL environment, enabling multi-turn autonomous exploration while strictly respecting configuration dependencies and validity constraints. \textbf{Rule-based reward functions for indirect feedback:} To cope with sparse performance feedback, we design reward functions that jointly enforce reasoning standardization, configuration correctness, and performance awareness, providing effective supervision without exhaustive benchmarking.
\textbf{A two-phase, data-efficient training pipeline:} We introduce a training pipeline that combines configuration-level learning with performance-driven exploration, enabling robust cross-scenario generalization with minimal retraining overhead. 
By integrating these innovations, TuneAgent learns structured and reliable tuning policies that seamlessly bridge abstract tuning targets with deployable kernel configurations.

We conduct extensive experiments across diverse tuning targets and real-world workloads. Results demonstrate that TuneAgent consistently outperforms existing baselines, delivering substantial performance improvements while maintaining high configuration validity. Additionally, TuneAgent exhibits strong scalability across real-world applications, achieving a 51.8\% performance boost for Nginx and 9.4\% for PostgreSQL, showcasing its practical deployment potential. Through its innovative kernel exploration mechanisms and training algorithms, TuneAgent paves the way for next-generation RL-driven OS optimization agents, fostering the development of more effective and adaptable Linux kernels.

%% file: Section/RelatedWork.tex
\section{Related Work}
\subsection{Linux Kernel Tuning}

Existing Linux kernel tuning methods can be broadly categorized into three classes.
\textbf{Heuristic tuning} relies on expert knowledge to manually adjust kernel configurations~\cite{braswell2008linux, Bovet2006UnderstandingLinuxKernel, Yi2014TuningLinuxKernel, ConfigFix}, which is effective in limited scenarios but labor-intensive and difficult to scale.
\textbf{ML-based tuning} applies data-driven optimization techniques~\cite{kernel4, kernel3, LAKE}, yet often depends on hand-crafted features and large labeled datasets, limiting generalization to unseen workloads.
\textbf{LLM-assisted tuning} leverages LLMs to reason about workload requirements~\cite{chenautoos, BYOS}, but struggles with efficient exploration and configuration validity in the presence of a large and constrained search space.
In contrast, \textbf{TuneAgent} introduces a rule-based RL framework that enables autonomous and valid kernel optimization with minimal labeled data, providing a scalable alternative to existing approaches.

\subsection{Reinforcement Learning in LLMs}
Recent studies show that RL can substantially enhance the reasoning capabilities of LLMs. DeepSeek-R1 \cite{DeepSeek-R1} demonstrated that RL without supervised fine-tuning (SFT) can induce emergent behaviors such as long chain-of-thought reasoning and self-correction. Light-R1 \cite{Light-R1} further combined SFT and RL via curriculum learning, achieving strong performance on mathematical reasoning. Search-R1 \cite{Search-R1} extends this paradigm with multi-turn interaction for improved question-answering. Building on this foundation, a series of domain-specific R1-style models, including Compiler-R1 \cite{Compiler-R1}, Fin‑R1 \cite{Fin-R1}, Surgery-R1 \cite{Surgery-R1}, Drive-R1 \cite{Drive-R1}, Scene-R1 \cite{Scene-R1}, MedVLM-R1 \cite{MedVLM-R1}, Ego-R1 \cite{Ego-R1}, GUI-R1 \cite{GUI-R1}, Light-R1 \cite{Light-R1}, LMM-R1 \cite{LMM-R1}, have shown that rule-driven RL can generalize across domains and give rise to specialized reasoning systems. To our knowledge, this paradigm has not yet been applied to OS-level optimization. TuneAgent bridges this gap by adapting rule-based RL to efficiently navigate the expansive kernel configuration space, enabling autonomous, effective, and deployable kernel tuning.

%% file: Section/Dataset.tex
\begin{figure*}[htbp]
\begin{center}
\centerline{\includegraphics[width=\textwidth]{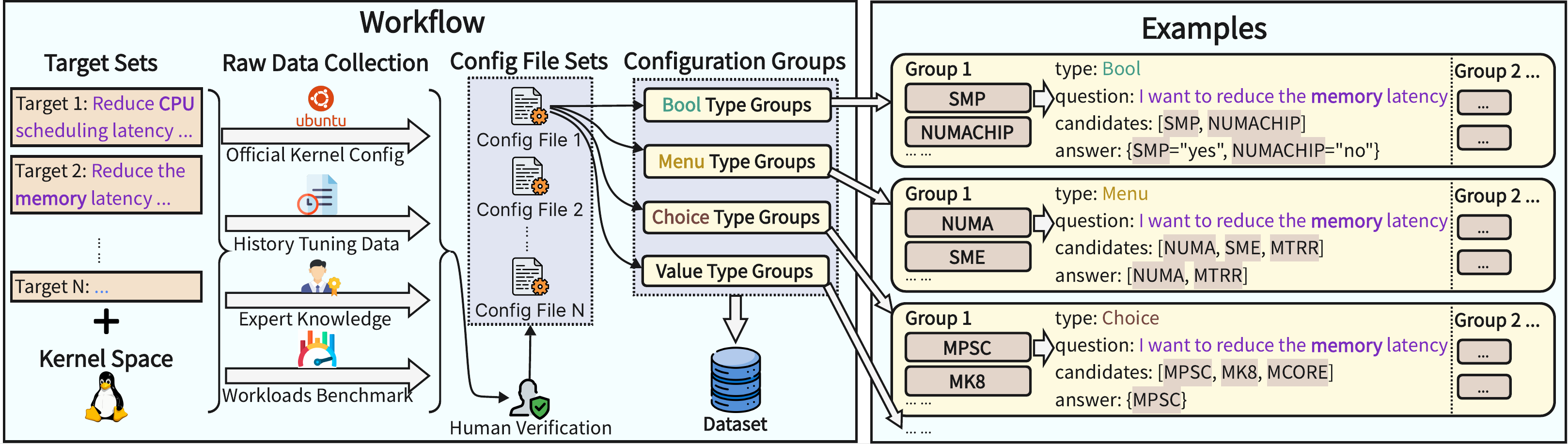}}
\caption{The dataset construction pipeline for TuneAgent. We first define a comprehensive set of tuning targets, then collect kernel configuration files from multiple complementary sources. After expert validation, configurations are organized and formatted into a unified schema, forming a structured configuration-group dataset for subsequent RL training.}
\Description{The dataset construction pipeline for TuneAgent.}
\label{fig:dataset}
\end{center}
\vskip -0.1in
\end{figure*}

\section{Dataset Construction}
\label{sec:data_construction}
Effective RL for kernel tuning critically depends on a high-quality, well-structured, and semantically consistent dataset that supports both efficient learning and robust generalization. This section presents our dataset construction pipeline, as illustrated in Figure~\ref{fig:dataset}.

\paragraph{Tuning Target Determination.}
Kernel tuning aims to optimize the performance of key kernel modules under diverse workloads. As summarized in Table~\ref{tab:targets} in the Appendix, we define a comprehensive set of expert-curated tuning targets spanning core kernel functionalities, including CPU scheduling, memory management, file I/O, process management, etc. For each subsystem, we specify representative tuning tasks and quantitative performance metrics that capture dominant system bottlenecks, such as latency, throughput, and resource utilization. By jointly covering fine-grained kernel mechanisms and end-to-end application scenarios, these targets establish a principled and comprehensive foundation for training and evaluating TuneAgent.

\paragraph{Raw Dataset Collection.} 
To ensure both coverage and realism, we collect kernel configuration data from four complementary sources: \textbf{(1) Official kernel configurations} extracted from official kernel documentation and widely used Linux distributions; \textbf{(2) Historical tuning data} accompanied by performance measurements (e.g., CPU utilization, I/O throughput, memory usage) obtained via profiling tools such as \textit{perf}; \textbf{(3) Expert tuning knowledge} derived from manual tuning logs and optimization best practices; \textbf{(4) Real-world workloads benchmarking} based on representative applications to simulate practical deployment scenarios. Together, these sources provide complementary perspectives that balance domain expertise, empirical evidence, and real-world applicability.

\paragraph{Dataset Preprocessing.} 
After data collection, we preprocess the raw configurations to construct a structured dataset suitable for RL training. Directly learning over the full kernel space—which contains more than 18{,}000 options governed by intricate constraints—is computationally and semantically infeasible. To address this challenge, we adopt a hierarchical, dependency-aware batching strategy that aggregates functionally related configurations into compact \emph{configuration groups}. Each group constitutes an independent training sample following a unified schema:
\begin{tcolorbox}[title={Example of a Configuration Group},
                  coltitle=white,
                  colback=blue!5,
                  colframe=blue!80!black,
                  fonttitle=\bfseries,
                  left=2mm,right=2mm,top=1mm,bottom=1mm,
                  boxrule=0.8pt]
\verb|"type"| \verb|:| \verb !Bool | Choice | Menu | Value! \\
\verb|"candidate"| \verb|:| \verb !CFG | [CFG-A, CFG-B, ...]! \\
\quad\verb|"question"| \verb|:| \verb !"Tuning Target"! \\
\quad\verb|"answer"| \verb|:| \verb !schema depends on <type>! 
\end{tcolorbox}
Here, \texttt{type} specifies the configuration category, \texttt{candidate} lists selectable options, \texttt{question} denotes the tuning objective, and \texttt{answer} defines the expected action. Depending on the configuration type, the answer schema is instantiated as: 
 \begin{itemize}
    \item \textbf{Bool}: enable or disable a configuration by \texttt{\{CFG="Yes/No"\}}
    \item \textbf{Menu}: activate one or more dependent configurations from a candidate set \texttt{[CFG-A, CFG-B, ...]}
    \item \textbf{Choice}: select exactly one configuration \texttt{\{CFG-A\}}
    \item \textbf{Value}: assign a valid parameter value \texttt{\{CFG="Literal"\}}
\end{itemize}

\paragraph{Automatic Configuration Generation.}
For each tuning target, we prompt LLMs with task-specific instructions to (1) enumerate only kernel configurations that are semantically relevant to the target and (2) assign their values in a step-by-step, unified, and structured workflow. During generation, kernel dependency rules and format constraints are explicitly enforced at prompt time, ensuring that all outputs are valid and well-formed by construction. Consequently, most generated configurations are already well-formed, and the remaining cases can be efficiently validated through dependency checks and human verification before RL training.

\paragraph{Resulting Dataset.}
The resulting dataset contains over 3{,}000 kernel configuration samples spanning a diverse set of tuning targets and kernel modules. All configurations are verified to successfully compile and boot, guaranteeing functional correctness and deployability. This expert-curated and structurally constrained dataset provides a reliable cold-start for RL, enabling data-efficient training, rapid convergence, and robust generalization across heterogeneous workloads. As a result, TuneAgent can be effectively deployed in real-world environments with minimal retraining overhead.

%% file: Section/Methdology.tex
\begin{figure*}[htbp]
\begin{center}
\centerline{\includegraphics[width=\textwidth]{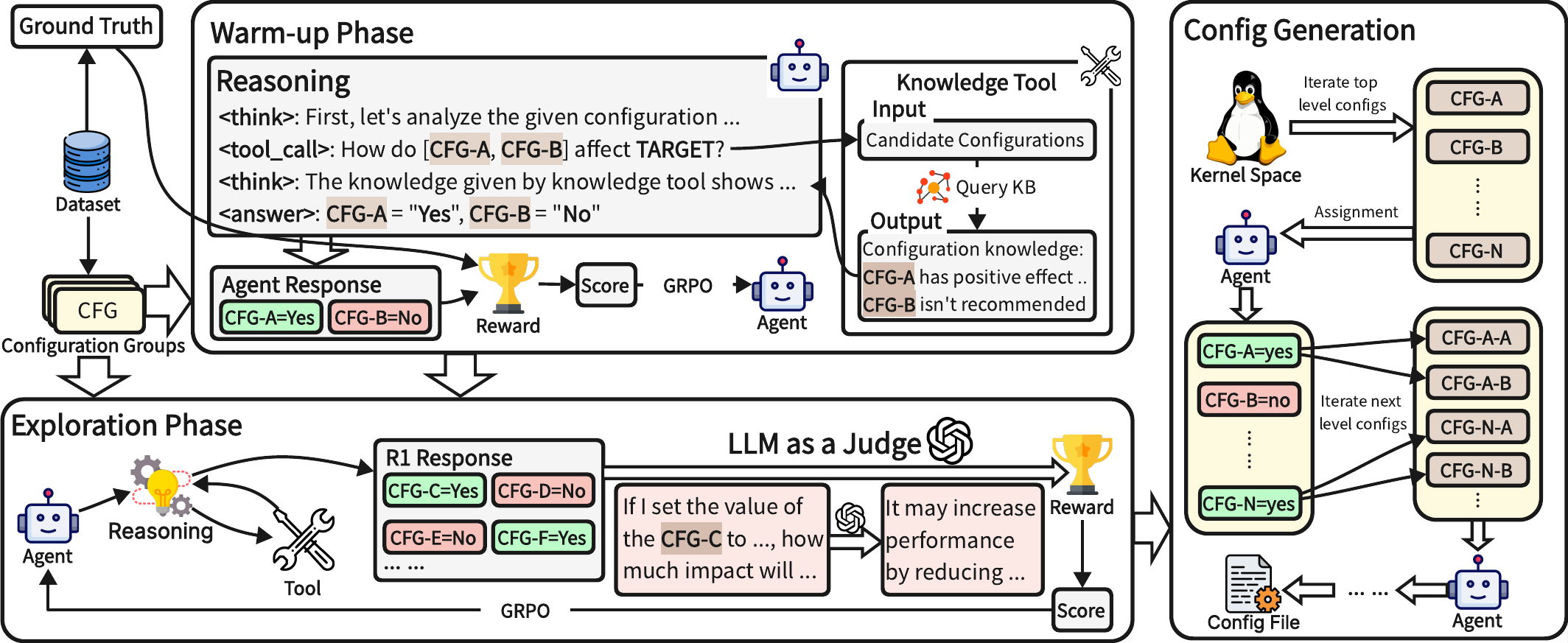}}
\caption{Overview of the TuneAgent framework. TuneAgent is trained via a two-phase pipeline: a \emph{Warm-up Phase} that enforces standardized reasoning and configuration correctness, followed by an \emph{Exploration Phase} that explicitly optimizes system performance through rule-based rewards and GRPO. The trained agent can then generate a complete, valid kernel configuration file for end-to-end tuning.}
\Description{Overview of the TuneAgent framework.}
\label{fig:framework}
\end{center}
\vskip -0.2in
\end{figure*}

\section{Methodology: TuneAgent}
In this section, we present the design of \textbf{TuneAgent}, an agentic framework for automated Linux kernel tuning. Figure~\ref{fig:framework} illustrates the overall training pipeline, which integrates a constrained RL environment, rule-based rewards, and a two-phase training strategy.

\subsection{Problem Formulation}
Linux kernel tuning aims to improve system performance by adjusting configurations under diverse workloads. Due to the hierarchical dependencies among kernel configurations and their non-linear impact on system behavior, we formulate \textbf{kernel tuning} as a sequential decision-making problem under the Markov Decision Process (MDP) paradigm~\cite{MDP}:
\begin{equation}
    \max_{\pi_{\theta}} \mathbb{E}_{\tau \sim \pi_{\theta}}\left[ \sum_{t=1}^{T} \gamma^t R(s_t, a_t, s_{t+1}) \right],
\end{equation}
where the components at time step $t$ are defined as follows:
\begin{itemize}
    \item \textbf{Kernel State} $\mathbf{s_t}$ represents the current configuration, workload characteristics, and runtime performance.
    \item \textbf{Action} $\mathbf{a_t}$ denotes a configuration modification, such as enabling, disabling, or assigning option values.
    \item \textbf{Reward Function} $\mathbf{R(s_t, a_t, s_{t+1})}$ evaluates both configuration validity and performance impact after executing $a_t$.
    \item \textbf{Discount Factor} $\gamma$ balances immediate performance gains against long-term tuning benefits.
    \item \textbf{Policy} $\pi_{\theta}$ selects actions conditioned on $\mathbf{s_t}$.
    \item \textbf{Trajectory} $\tau$ denotes the sequence of state-action transitions induced by $\pi_{\theta}$ during tuning.
\end{itemize}
This formulation enables TuneAgent to learn tuning strategies through iterative agent–environment interaction that improve system performance while preserving configuration validity.

\subsection{Environment and Agent Setup}
\paragraph{Kernel Configuration Environment Construction.}
To enable agent-driven kernel tuning, we abstract the intricate kernel configuration space $C$ into a constraint-aware interactive environment $\mathcal{E}$ suitable for RL. Functionally related options aligned with tuning objectives are organized into groups $G_i$ to reduce exploration complexity while preserving structural dependencies:
\begin{equation}
\mathcal{E} = \mathcal{F}(C) = \{G_1,\dots,G_K\}, \quad G_k \subset C,
\end{equation}
At step $t$, the agent performs actions restricted to the active group, i.e., $a_t \in \mathcal{A}(s_t, G_{k_t})$, where $\mathcal{A}(s_t, G_{k_t})$ denotes dependency-consistent modifications under state $s_t$. The environment then updates the system state through a constraint-aware transition:
\begin{equation}
s_{t+1} = \mathcal{E}(s_t, a_t) = \Pi_{\mathrm{valid}}\!\left(s_t \oplus a_t\right),
\end{equation}
where $\oplus$ applies configuration updates and $\Pi_{\mathrm{valid}}$ enforces dependency constraints to ensure configuration validity. This abstraction transforms the kernel space into a structured environment that enables efficient exploration while preserving system correctness.

\paragraph{Agent Action Space Initialization.}
In TuneAgent, each action $a_t = \big(a_t^{\mathrm{grp}},\, a_t^{\mathrm{cfg}}\big) \in \mathcal{A}$ is decomposed into \emph{configuration group selection} $a_t^{\mathrm{grp}}$ and \emph{configuration assignment} $a_t^{\mathrm{cfg}}$ within that group. The corresponding action log-likelihood factorizes as:
\begin{equation}
\log \pi(a_t \mid s_t)
=
\log \pi(a_t^{\mathrm{grp}} \mid s_t)
+
\log \pi(a_t^{\mathrm{cfg}} \mid s_t, a_t^{\mathrm{grp}}).
\end{equation}
Operationally, at each step the agent first determines which configuration region to tune next, and then generates dependency-consistent updates within that region, reducing exploration complexity while ensuring valid configuration transitions.

\subsection{Rule-based Reward Design}
\label{sec:train_algo_design}
Kernel tuning requires exploring a highly constrained configuration space, where improper exploration or reward design can lead to excessive invalid trials and costly evaluations. To address this, we design a \emph{tuning-oriented reward function} that integrates format standardization, configuration correctness, and performance awareness to enable efficient and reliable policy optimization.

\paragraph{Format Reward ($R_{format}$)} 
To promote structured decision-making, the agent is encouraged to follow a predefined response format: reasoning within \texttt{<think>} tags, tool invocation within \texttt{<tool\_call>} tags, and configuration updates within \texttt{<answer>} tags. The format reward is defined as:
\begin{equation}
    R_{format} =
    \begin{cases}
        1 & \textit{if the format matches constraints} \\
        0 & \textit{otherwise}
    \end{cases}
\end{equation}
Here, \texttt{<tool\_call>} queries a domain knowledge base built from kernel documentation, historical tuning data, and expert knowledge, providing reliable tuning and interpretable guidance.

\paragraph{Answer Reward ($R_{answer}$)} 
Kernel configurations follow strict type semantics, and invalid modifications may degrade performance or cause system crashes. To ensure safe actions, we design type-aware rewards for four configuration categories:
\begin{itemize}
    \item \textbf{Bool}: reward is given for assigning a valid ``yes'' or ``no'' value that respects configuration dependencies, ensuring the option is correctly enabled or disabled.
    \item \textbf{Menu}: reward is given when selecting one or more valid options from the candidate set, guaranteeing coherent activation of related configurations.
    \item \textbf{Choice}: reward is given when exactly one valid option is selected from mutually exclusive candidates, preserving the exclusivity constraint of the configuration group.
    \item \textbf{Value}: reward is given when the assigned value falls within the permitted domain and satisfies range or type constraints, ensuring safe and executable parameter settings.
\end{itemize}
A score of 1 is given to valid modifications and 0 otherwise, ensuring syntactic and operational correctness.

\paragraph{Performance Reward ($R_{perf}$)} 
Evaluating tuned kernels traditionally requires recompilation, deployment, and benchmarking, resulting in delayed feedback and inefficient RL exploration. To provide timely performance guidance, we adopt an LLM-as-a-Judge framework \cite{judge} to approximate performance evaluation during training. During training, the judge provides a low-cost proxy estimate of the expected performance impact based on configuration semantics and available profiling evidence, enabling low-cost and consistent feedback without repeated system-level benchmarking:
\begin{equation}
    R_{\text{perf}} = \sum_{i=1}^{N} \left( \frac{P_{\text{new}, i} - P_{\text{base}, i}}{P_{\text{base}, i}} \right) \cdot \left( 1 + \lambda_i \cdot \frac{C_{\text{config}, i}}{C_{\text{max}, i}} \right)
\end{equation}
where \( N \) is the number of configuration modifications, \( P_{\text{new}, i} \) and \( P_{\text{base}, i} \) are the new and baseline performance scores for modification \( i \), \( \lambda_i \), \( C_{\text{config}, i} \) and \( C_{\text{max}, i} \) are related to modification complexities. This formulation encourages meaningful performance gains while discouraging unnecessary environment exploration, substantially accelerating training.

\subsection{Reinforcement Learning in Kernel Tuning}
\label{sec:rl_kernel_tuning}
To enable efficient and stable training, we propose a \emph{kernel tuning–oriented RL framework}, where TuneAgent is trained through a unified Group Relative Policy Optimization (GRPO)~\cite{GRPO} pipeline with \emph{two complementary phases serving different objectives}.

\paragraph{Phase I: Standardization-oriented Warm-up}
The warm-up phase aims to establish configuration-level correctness and structural behaviors. Policy learning is guided exclusively by $R_{format}$ and $R_{answer}$, encouraging the agent to (i) follow structured reasoning and output formats, (ii) invoke external knowledge tools appropriately, and (iii) handle different configuration types correctly. Starting from an initial policy $\pi_{\theta_0}$ and kernel state $s_t$, a group of $G$ candidate actions is sampled as:
\begin{equation}
    a_t^i \sim \pi_\theta(a_t^i \mid s_t), \quad i \in \{1, \ldots, G\},
\end{equation}
and executed to generate multi-step interaction trajectories:
\begin{equation}
    \tau = \{(s_t, a_t, r_t, s_{t+1})\}_{t=1}^{T},
\end{equation}
Group-based sampling enables relative comparison among candidate actions under the same state, providing a stable initialization before performance-driven exploration.

\paragraph{Phase II: Performance-aware Exploration}
The exploration phase shifts from structural correctness to system-level performance awareness. In this phase, $R_{perf}$ is introduced to guide the agent toward configurations that yield measurable performance improvements while preserving validity learned in Phase I. To mitigate noisy and workload-dependent reward scales, rewards within each group are normalized to compute relative advantages:
\begin{equation}
    A_i = \frac{r_i - \mu}{\sigma}, \quad
    \mu = \frac{1}{G} \sum_{i=1}^{G} r_i, \quad
    \sigma = \sqrt{\frac{1}{G} \sum_{i=1}^{G} (r_i - \mu)^2}.
\end{equation}
The policy is then updated using a clipped GRPO objective, progressively refining the agent’s decisions toward configurations with higher relative performance gains:
\begin{equation}
    \mathcal{L}(\theta) =
    \mathbb{E}_{s_t, a_t^i \sim \pi_\theta}
    \left[
        \min \left(
            \frac{\pi_\theta(a_t^i \mid s_t)}{\pi_{\theta_{\text{old}}}(a_t^i \mid s_t)},
            1 + \epsilon
        \right)
        A_i
    \right],
\end{equation}
Overall, this GRPO-based optimization allows TuneAgent to progressively shift from rule-guided behavior learned during warm-up to performance-driven exploration, achieving efficient and robust kernel tuning while maintaining configuration validity.

\begin{tcolorbox}[
  colback=cyan!5!white,     
  colframe=teal!90!black,   
  coltitle=white,           
  title=Kernel Tuning Assistant Template,  
  fonttitle=\bfseries,      
  sharp corners=south,      
  boxrule=1pt               
]
You are a kernel tuning assistant. You may first analyze the \textbf{tuning target}, and then explore the \textbf{kernel space} to provide the corresponding tuning decisions. The reasoning process and the final decision should be enclosed within \textcolor{blue}{\texttt{<think>}} and \textcolor{magenta}{\texttt{<answer>}} tags, respectively, i.e., \textcolor{blue}{\texttt{<think>}} reasoning here \textcolor{blue}{\texttt{</think>}} \textcolor{magenta}{\texttt{<answer>}} decision here \textcolor{magenta}{\texttt{</answer>}}. If needed, use \textcolor{brown}{\texttt{<tool\_call>}}...\textcolor{brown}{\texttt{</tool\_call>}} to retrieve additional knowledge or perform analysis.
\end{tcolorbox}

%% file: Section/Experiment.tex
\begin{table*}[ht]
\caption{Performance comparison of TuneAgent and representative baselines across key kernel modules and overall system performance. The best UnixBench result in each column is marked with \textbf{bold} $*$ and the second-best with \textit{italic} $\dagger$.}
\centering
\begin{tabular}{lccccccccl}
\toprule
\multirow{2}{*}{\textbf{Model}} & \multicolumn{7}{c}{\textbf{Kernel Module}} & \multirow{2}{*}{\textbf{Overall}} \\
 & \textbf{CPU} & \textbf{Memory} & \textbf{File} & \textbf{Pipe} & \textbf{Shell} & \textbf{System Call} & \textbf{Process} & \\
\cmidrule(lr){2-8}
Heuristic (Default)        & 3074.4 & 316.5 & 1123.9 & 427.9 & 1091.5 & 178.8 & 187.3 & 627.2 \\
o3-mini                    & 3213.4 & 333.9 & 1176.1 & 471.5 & 789.2 & 179.5 & 189.3 & 598.3\down{28.9} \\
DeepSeek-R1                & 3200.3 & 337.1 & 1047.2 & 464.9 & 1083.9 & 256.6 & 189.0 & \textit{650.5} $\dagger$\up{23.3} \\
GPT-4o                     & 3264.1 & 326.0 & 1201.8 & 512.3 & 1051.7 & 194.6 & 201.1 & 632.9\up{5.7} \\
GPT-4o-mini                & 3147.9 & 323.1 & 1173.5 & 509.9 & 1036.9 & 190.4 & 182.4 & 631.8\up{4.6} \\
AutoOS                     & 3315.7 & 290.8 & 1166.5 & 439.4 & 1022.4 & 208.3 & 193.8 & 638.8\up{11.6} \\
Qwen2.5-3B-Instruct        & 3211.7 & 317.7 & 1143.6 & 457.7 & 1011.0 & 189.0 & 183.8 & 615.4\down{11.8} \\
Qwen2.5-7B-Instruct        & 3318.8 & 329.2 & 1185.8 & 494.1 & 1044.9 & 207.9 & 190.5 & 619.6\down{7.6} \\
\textbf{TuneAgent-3B (Ours)}            & 3296.9 & 328.8 & 1184.2 & 489.0 & 1036.5 & 204.3 & 188.2 & 643.8\up{16.6} \\
\textbf{TuneAgent-7B (Ours)}            & 3331.5 & 326.2 & 1209.3 & 505.7 & 1044.5 & 209.2 & 199.6 & \textbf{662.2} $*$\textbf{\up{35.0}} \\
\bottomrule
\end{tabular}
\label{tab:main}
\end{table*}

\section{Experiment}
We evaluate the effectiveness, efficiency, and generalizability of TuneAgent through a comprehensive set of empirical studies designed to answer the following research questions (RQs): \textbf{RQ1.} How does TuneAgent compare with existing baselines in kernel tuning performance? \textbf{RQ2.} Can TuneAgent consistently generate valid, correct, and bootable kernel configurations? \textbf{RQ3.} How do different reward components in TuneAgent affect tuning performance and configuration validity? \textbf{RQ4.} How well does TuneAgent generalize to diverse real-world applications? \textbf{RQ5.} How do different reward components influence TuneAgent’s training efficiency and convergence? \textbf{RQ6.} How does TuneAgent achieve fine-grained and reliable kernel tuning in practice?

\subsection{Experimental Setup}
\paragraph{Models.}
We utilize \textbf{Qwen2.5-3B-Instruct} and \textbf{Qwen2.5-7B-Instruct}~\cite{qwen2} as the base language models for TuneAgent to evaluate its effectiveness across different model scales.

\paragraph{Baselines.}
We compare TuneAgent with three categories of baselines:
(1) \textbf{Heuristic Tuning}, which uses the default kernel configuration as a proxy for expert-crafted settings commonly adopted in practice;
(2) \textbf{Vanilla LLMs}, where models such as GPT-4o and DeepSeek-R1 generate configurations via few-shot prompting with chain-of-thought reasoning;
and (3) \textbf{LLM-Assisted Tuning}, represented by AutoOS~\cite{chenautoos}, a state machine–based framework for iterative kernel exploration.
Notably, ML-based methods are excluded due to the lack of publicly reproducible implementations.

\paragraph{Benchmark.}
We adopt UnixBench~\cite{unixbench} to evaluate kernel performance. It evaluates multiple kernel modules (e.g., CPU, Memory, File I/O) via standardized sub-tests and aggregates their results into an overall system performance score, enabling consistent and quantitative comparison across methods.

\paragraph{Implementation.}
All methods are evaluated under identical hardware and software environments, with multiple independent runs per tuning target to reduce randomness. GPT-4o-mini is used for knowledge base construction and LLM-as-a-Judge during training. LLM-assisted baselines are executed on four NVIDIA H100 GPUs.

\subsection{Main Results (RQ1)}
Table~\ref{tab:main} summarizes the performance comparison between TuneAgent and representative baselines across kernel modules as well as the overall system. We highlight the following key observations.

\paragraph{TuneAgent consistently improves kernel tuning performance.}
Both \textbf{TuneAgent-3B} and \textbf{TuneAgent-7B} consistently outperform their corresponding base models (\textbf{Qwen2.5-3B-Instruct} and \textbf{Qwen2.5-7B-Instruct}) at both the module and system levels. In particular, TuneAgent-7B attains the highest overall performance score, while TuneAgent-3B also substantially outperforms most baselines. These results demonstrate the effectiveness of combining RL with structured, constraint-aware kernel interactions.

\paragraph{Reinforcement learning is critical for effective kernel tuning.}
Models trained with RL exhibit clear advantages in system-level performance. TuneAgent-7B attains the best overall result, while DeepSeek-R1 (also RL-trained) ranks second. Notably, TuneAgent-3B performs competitively with, and in some cases surpasses, much larger models such as GPT-4o. This suggests that RL plays a crucial role in aligning agent behavior with tuning targets and enabling efficient exploration of the constrained kernel space.

\paragraph{Performance varies across kernel modules due to data coverage and model capacity.}
Although TuneAgent-7B achieves the strongest overall performance, its gains are not uniformly dominant across all modules (e.g., \emph{Memory}, \emph{Pipe}, and \emph{Shell}). This variation reflects differences in training data coverage and the inherent complexity of modeling fine-grained kernel behavior. Larger general-purpose models (e.g., DeepSeek-R1, GPT-4o) also perform strongly on certain modules, likely benefiting from broader pretraining. 

\paragraph{Summary.}
Overall, these results indicate that effective kernel tuning depends jointly on an appropriate learning paradigm, sufficient domain-specific data coverage, and adequate model capacity. RL enables better alignment with system-level objectives and more efficient exploration of the constrained kernel space, while structured, validity-aware interactions further enhance reliability. At the same time, broader and more balanced data coverage is essential for consistent gains across modules, and larger models can better capture complex configuration interdependencies. 


\begin{figure*}[htbp]
\centering
\includegraphics[width=\textwidth]{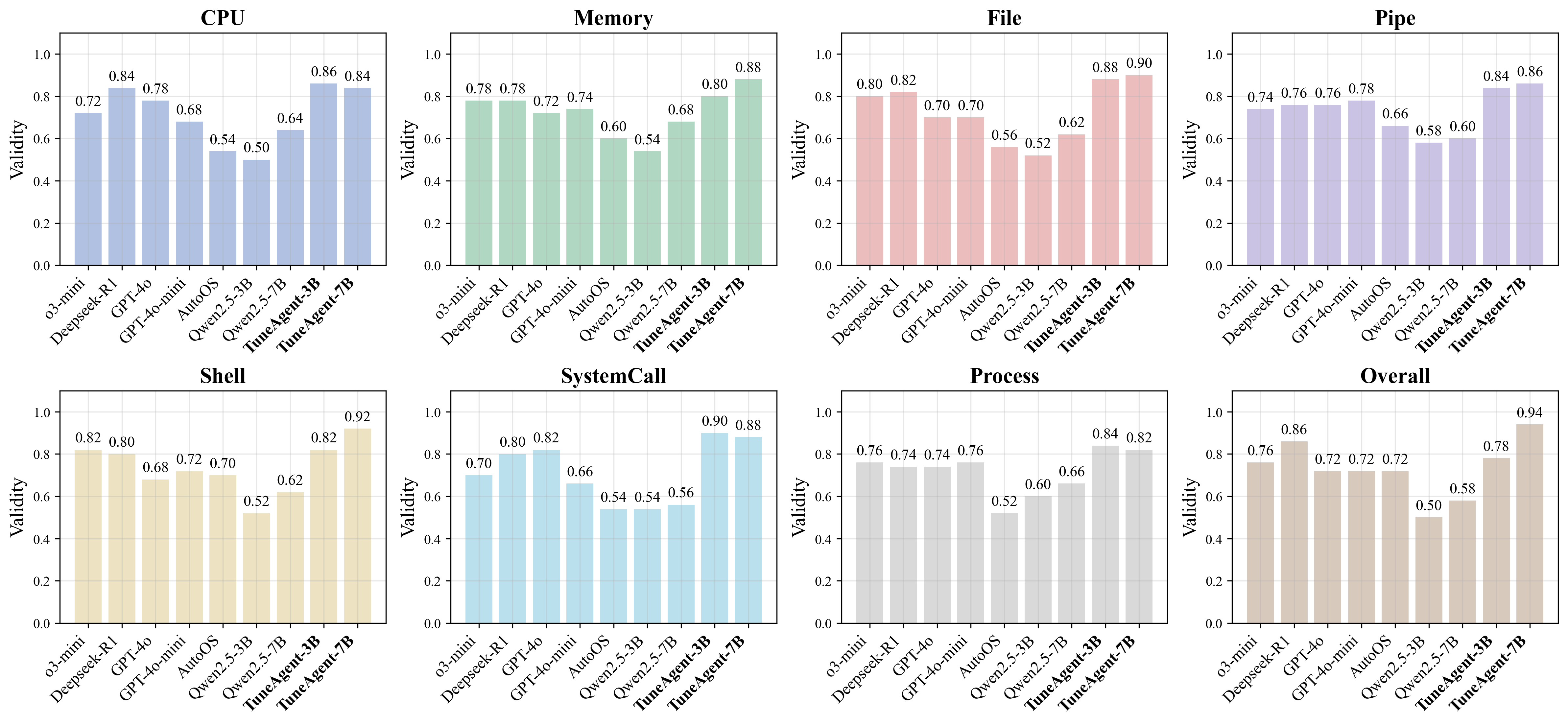}
\caption{Configuration validity comparison across different kernel tuning targets. For each target, 50 kernel configurations are generated by each method and evaluated based on whether the resulting kernel can successfully compile and boot.}
\Description{Configuration validity comparison across different kernel tuning targets. 
For each target, 50 kernel configurations are generated by each method and evaluated based on whether the resulting kernel can successfully compile and boot.}
\label{fig:validity}
\vspace{-0.8\baselineskip}
\end{figure*}

\subsection{Configuration Validity Analysis (RQ2)}
Figure~\ref{fig:validity} reports the configuration validity of TuneAgent and baselines across different tuning targets. For each target, we generate 50 kernel configurations per method and evaluate validity based on whether the resulting kernel can successfully compile and boot. Validity is measured as the proportion of configurations that are functionally deployable in practice.

\paragraph{TuneAgent achieves consistently higher validity.}
Both \textbf{TuneAgent-3B} and \textbf{TuneAgent-7B} achieve the highest or near-highest validity across all tuning targets. This demonstrates TuneAgent’s ability to reliably generate executable and bootable configurations under heterogeneous scenarios, rather than producing ad hoc modifications.

\paragraph{Vanilla LLMs show unstable validity.}
In contrast, vanilla LLM baselines such as GPT-4o exhibit lower and more variable success rates, especially on configuration-sensitive targets (e.g., \emph{Shell}, \emph{System Call}, and \emph{Process}). This instability indicates that prompt-based generation alone is insufficient to ensure correctness in highly constrained kernel configuration spaces.

\paragraph{Structured RL is crucial for correctness.}
While DeepSeek-R1 benefits from RL and outperforms vanilla LLMs in validity, it remains consistently inferior to TuneAgent on more complex modules. This gap highlights the importance of TuneAgent’s structured interaction design and validity-aware rewards, which explicitly enforce configuration semantics and dependency constraints.

\paragraph{Implications for practical deployment.}
Overall, TuneAgent significantly outperforms all baselines in aggregated validity, demonstrating that it not only improves tuning performance (RQ1) but also ensures the practical deployability of generated configurations, making it suitable for real-world kernel tuning scenarios.

\vspace{-0.6\baselineskip}
\begin{table}[htbp]
\centering
\caption{Impact of reward schemes on TuneAgent optimization performance and configuration validity.}
\label{tab:ablation}
\begin{tabular}{c|cc}
\toprule
\textbf{Reward Scheme} & \textbf{Performance} & \textbf{Validity} \\
\midrule
Qwen-7B                                 & 619.6 & 58.4\% \\
\midrule
$R_{format}$                            & 643.8\up{24.2} & 62.2\%\up{3.8\%} \\
$R_{format}$ \& $R_{perf}$              & 654.3\up{34.7} & 63.3\%\up{4.9\%} \\
$R_{format}$ \& $R_{answer}$            & 644.4\up{24.8} & 78.7\%\up{20.3\%} \\
\textbf{TuneAgent-7B}                   & \textbf{662.2}\up{42.6} & \textbf{93.8}\%\up{35.4\%} \\
\bottomrule
\end{tabular}
\vspace{-0.8\baselineskip}
\end{table}

\subsection{Ablation Study (RQ3)}
We conduct an ablation study to evaluate the contributions of different reward components in TuneAgent, focusing on system-level performance improvement and configuration validity.

\paragraph{Impact on optimization performance.}
As shown in Table~\ref{tab:ablation}, $R_{format}$ alone already improves performance over the base model (Qwen-7B), indicating that enforcing structured outputs facilitates more effective tuning. Further incorporating $R_{perf}$ yields additional gains, demonstrating the importance of performance-aware rewards for guiding system-level optimization under sparse feedback.

\paragraph{Impact on configuration validity.}
Reward design also has a clear effect on configuration validity. While $R_{format}$ and $R_{format}$+$R_{perf}$ result in only marginal validity improvements, adding $R_{answer}$ substantially increases the proportion of valid configurations, highlighting the importance of type- and constraint-aware rewards. 

\paragraph{Summary.}
Overall, the full reward scheme achieves the best trade-off between performance and validity, consistently outperforming all partial variants and demonstrating the effectiveness of complementary reward design for kernel tuning.

\begin{figure}[htbp]
\centering
\includegraphics[width=80mm]{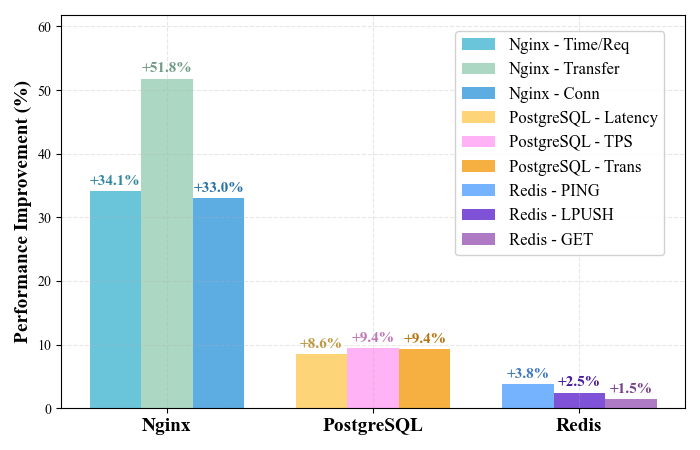}
\caption{Performance improvements achieved by TuneAgent on Nginx, PostgreSQL, and Redis across multiple metrics.}
\Description{Performance improvement of TuneAgent across various system load applications.}
\label{fig:app}
\end{figure}

\subsection{Generalization Performance Analysis (RQ4)}
To assess the generalizability of TuneAgent, we evaluate its performance on three real-world workloads—\texttt{Nginx}, \texttt{PostgreSQL}, and \texttt{Redis}—spanning I/O-bound, CPU-intensive, and memory-intensive scenarios. Performance is measured as the percentage improvement over the default kernel configuration using ApacheBench~\cite{apache_bench}, Sysbench~\cite{sysbench}, and Redis Benchmark~\cite{redis_benchmarks}, respectively.

\paragraph{Result Analysis.}
As shown in Figure~\ref{fig:app}, TuneAgent consistently improves performance across all applications. On \texttt{Nginx}, it achieves substantial gains of up to \textbf{51.8\%}. For \texttt{PostgreSQL}, TuneAgent yields stable improvements of approximately \textbf{8.6\%--9.4\%} across latency and throughput metrics. On the highly optimized \texttt{Redis}, TuneAgent still delivers consistent, though smaller, gains (\textbf{1.5\%--3.8\%}), highlighting its robustness in performance saturated environments.

\paragraph{Summary.}
These results show that TuneAgent generalizes well to diverse real-world systems without workload-specific retraining, highlighting its scalability for practical kernel tuning.

\begin{figure}[htbp]
\centering
\includegraphics[width=80mm]{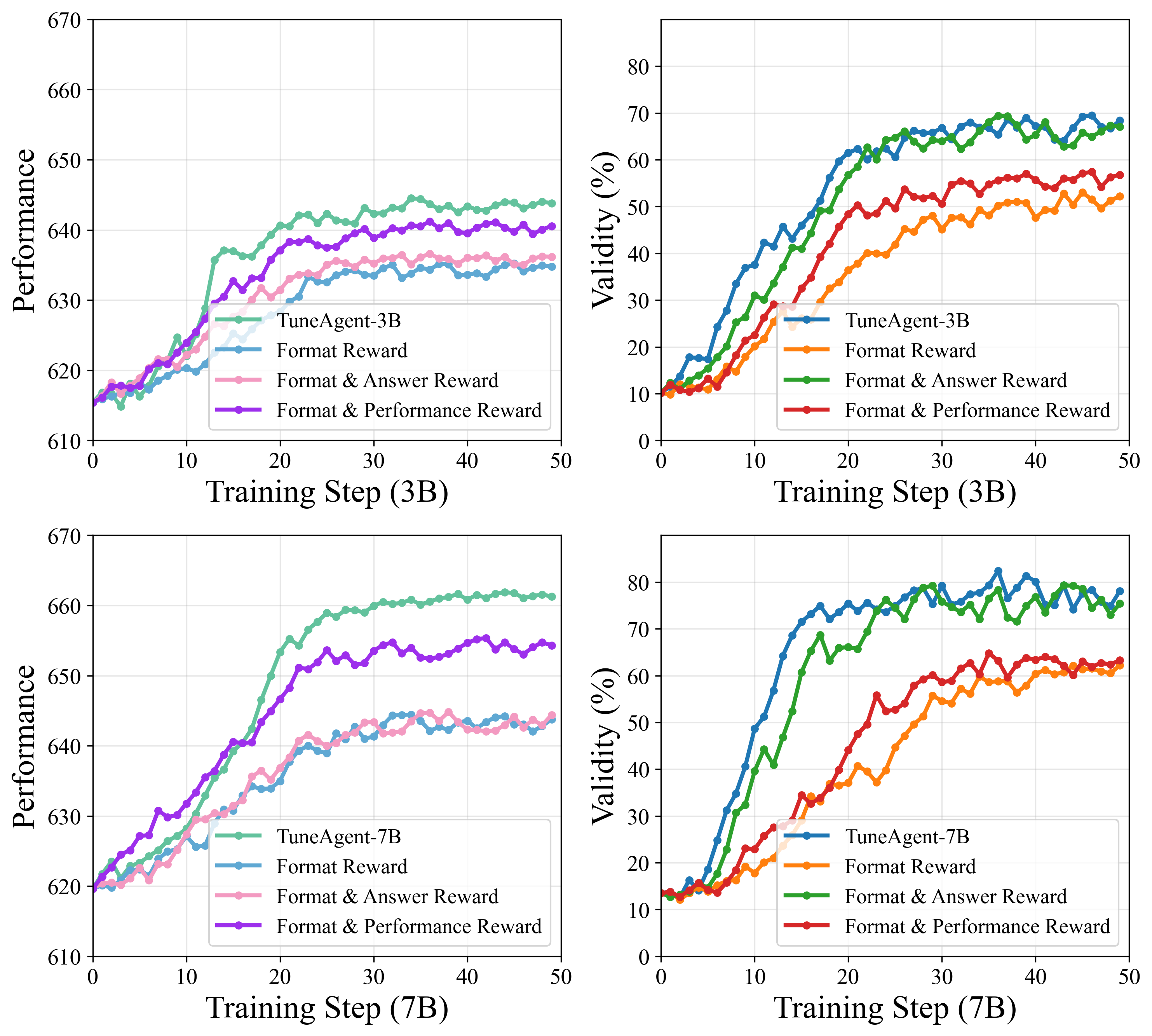}
\caption{Impact of different reward schemes on optimization performance and configuration validity during training across models with different parameter scales.}
\Description{Comparison of overall rewards for TuneAgent-7B and TuneAgent-3B models during training.}
\label{fig:ablation}
\end{figure}

\subsection{Training Efficiency Analysis (RQ5)}
We analyze how different reward components affect TuneAgent’s performance, configuration validity, and training efficiency.

\paragraph{Effect on convergence behavior.}
As shown in Figure~\ref{fig:ablation}, reward design significantly influences convergence behavior. Performance-aware rewards lead to faster and more stable convergence, indicating more efficient exploration of the kernel space, while removing them results in slower convergence and higher training variance.

\paragraph{Role of validity-aware rewards.}
Validity-aware rewards are essential for ensuring correctness during training. Without explicit validity or answer constraints, models frequently generate configurations that violate kernel dependencies or fail to compile, even if short-term performance gains are observed.

\paragraph{Overall training efficiency.}
By jointly integrating format consistency, configuration validity, and performance awareness, the full reward formulation achieves the best balance between training efficiency, correctness, and optimization performance, enabling stable and reliable kernel tuning.

\subsection{Case Study (RQ6)}
We examine how TuneAgent achieves reliable and fine-grained kernel tuning in practice, outperforming the base model (Qwen2.5) and, in some cases, substantially larger LLM-based approaches.

\paragraph{Early-stage instability without structured outputs.}
At early training stages, the base model fails to produce standardized outputs, leading to incorrect and unstable modifications with poor validity and performance (e.g., disabling \texttt{CONFIG\_FILE\_LOCKING} or overly small \texttt{CONFIG\_FRAME\_WARN} values).

\paragraph{Target-aware and fine-grained configuration reasoning.}
After learning structured outputs, TuneAgent evaluates the relevance and impact of options with respect to the tuning target. With sufficient training data, it explores fine-grained options that are often overlooked by larger models, such as adjusting tracer-related sub-options of \texttt{CONFIG\_FTRACE} instead of making coarse-grained enable/disable decisions.

\paragraph{Balancing performance and reliability.}
This fine-grained, target-aware tuning enables TuneAgent to improve performance while minimizing functional side effects, demonstrating its practical advantage in balancing optimization effectiveness and reliability.

%% file: Section/Conclusion.tex
\section{Conclusion}
We present \textbf{TuneAgent}, an RL-based framework for automated Linux kernel tuning that integrates rule-based rewards with tool-augmented reasoning. TuneAgent employs a two-phase training strategy to improve training efficiency and accelerate convergence. Extensive experiments show that TuneAgent outperforms existing approaches with substantial performance gains while maintaining high configuration validity. Moreover, TuneAgent generalizes well across diverse real-world workloads, highlighting its scalability and practicality for efficient and reliable system optimization.

%% file: Section/Appendix.tex
\appendix

\section*{Appendix}
In the appendix, we provide additional materials that cannot fit into the main manuscript due to page limit, including pseudocode, experimental settings, and additional results.

\begin{algorithm}[htbp]
\caption{TuneAgent Kernel Tuning Algorithm}
\label{alg:tune}
\begin{algorithmic}[1]
\REQUIRE Dataset $\mathcal{D}$, base policy $\pi_{\theta_0}$, knowledge base $\mathrm{KB}$, reward weights $\alpha$, $\beta$, $\eta$, clipping parameter $\epsilon_{\mathrm{clip}}$, group size $G$
\ENSURE Optimized policy $\pi_\theta$ and tuned kernel configurations

\STATE Initialize policy $\pi_\theta \leftarrow \pi_{\theta_0}$ with the base LLM parameters
\STATE Construct a constraint-aware kernel environment $\mathcal{E}$ from $\mathcal{D}$
\STATE Construct configuration groups $\{G_k\}_{k=1}^{K}$ from $\mathcal{D}$

\FOR{each training phase $\phi \in \{\textsc{Warm-up}, \textsc{Exploration}\}$}
    \FOR{each tuning target and configuration group $G_k$}
        \STATE Observe the current kernel state $s_t$ from $\mathcal{E}$
        \STATE Sample $G$ candidate responses from the current policy:
        \[
        a_t^{i} \sim \pi_\theta(\cdot \mid s_t, G_k), \quad i = 1,\ldots,G
        \]

        \FOR{each candidate action $a_t^{i}$}
            \STATE Optionally query $\mathrm{KB}$ through \texttt{<tool\_call>} for domain-specific tuning knowledge
            \STATE Execute or simulate $a_t^{i}$ in the constraint-aware environment and obtain $s_{t+1}^{i}$
            \STATE Compute the format reward:
            \[
            R_{\mathrm{format}}^{i} = R_{\mathrm{format}}(s_t, a_t^{i})
            \]
            \STATE Compute the answer reward:
            \[
            R_{\mathrm{answer}}^{i} = R_{\mathrm{answer}}(s_t, a_t^{i})
            \]
            \IF{$\phi = \textsc{Exploration}$}
                \STATE Compute the performance-aware reward:
                \[
                R_{\mathrm{perf}}^{i} = R_{\mathrm{perf}}(s_t, a_t^{i}, s_{t+1}^{i})
                \]
            \ELSE
                \STATE Set $R_{\mathrm{perf}}^{i} = 0$
            \ENDIF
            \STATE Combine reward components:
            \[
            R^{i} =
            \alpha R_{\mathrm{answer}}^{i}
            + \beta R_{\mathrm{format}}^{i}
            + \eta R_{\mathrm{perf}}^{i}
            \]
        \ENDFOR

        \STATE Normalize rewards within the candidate group:
        \[
        \mu = \frac{1}{G}\sum_{i=1}^{G} R^{i},
        \quad
        \sigma = \sqrt{\frac{1}{G}\sum_{i=1}^{G}(R^{i}-\mu)^2}
        \]
        \STATE Compute relative advantages:
        \[
        A^{i} = \frac{R^{i}-\mu}{\sigma + \delta},
        \quad i=1,\ldots,G
        \]
        where $\delta$ is a small constant for numerical stability.

        \STATE Update the policy using the clipped GRPO objective:
        \[
        \mathcal{L}(\theta)
        =
        \mathbb{E}_{i}
        \left[
        \min
        \left(
        \rho_i(\theta) A^{i},
        \mathrm{clip}
        \left(
        \rho_i(\theta),
        1-\epsilon_{\mathrm{clip}},
        1+\epsilon_{\mathrm{clip}}
        \right) A^{i}
        \right)
        \right],
        \]
        where
        \[
        \rho_i(\theta)
        =
        \frac{
        \pi_\theta(a_t^{i}\mid s_t, G_k)
        }{
        \pi_{\theta_{\mathrm{old}}}(a_t^{i}\mid s_t, G_k)
        }.
        \]
    \ENDFOR
\ENDFOR

\STATE Generate final configurations using the optimized policy $\pi_\theta$
\STATE Validate dependency consistency, compilation, and bootability of the generated configurations
\RETURN Optimized kernel configurations
\end{algorithmic}
\end{algorithm}



\section{Algorithm Overview}
Algorithm~\ref{alg:tune} summarizes the end-to-end kernel tuning procedure of TuneAgent. The agent interacts with a constraint-aware kernel environment in an episodic loop: at each step, it selects an action based on the current policy and state, optionally queries a knowledge base, executes the action, and receives a composite reward integrating format consistency, configuration validity, and performance awareness.



\begin{figure*}[htbp]
\begin{center}
\centerline{\includegraphics[width=\textwidth]{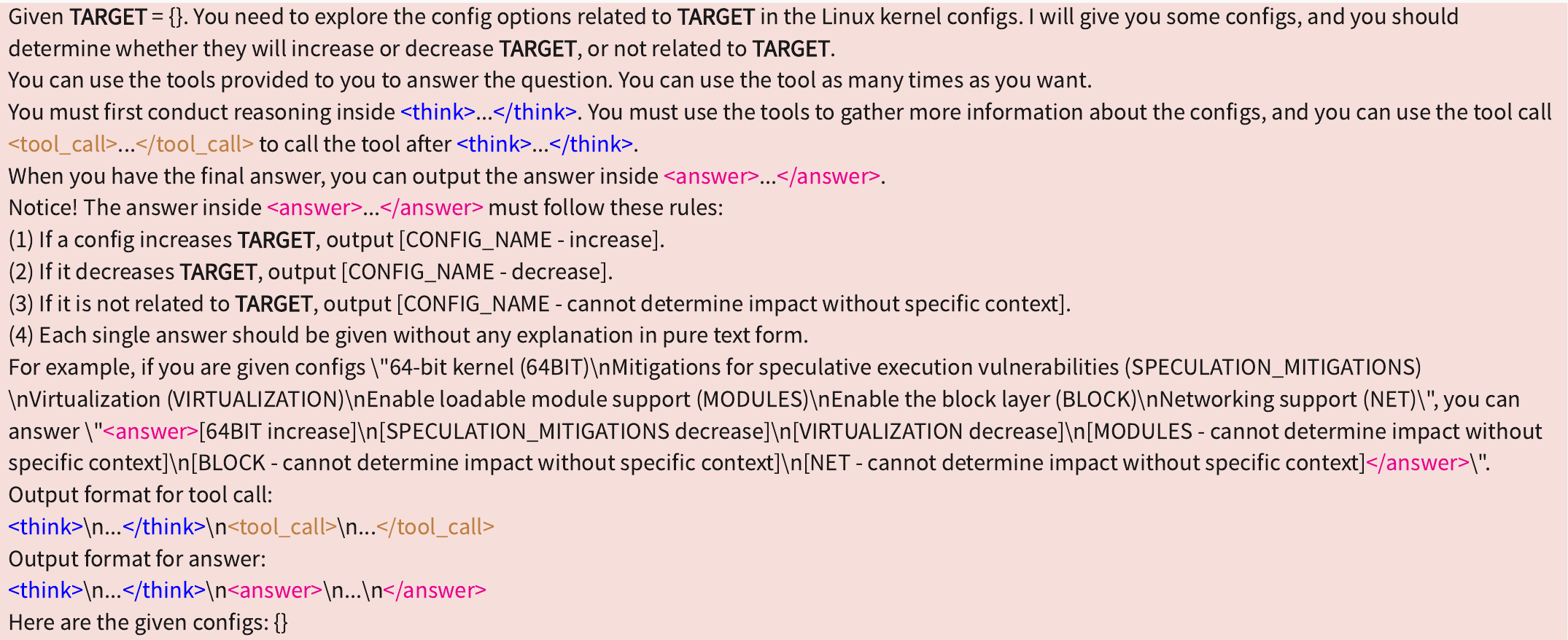}}
\caption{Prompt for bool type configurations.}
\Description{Prompt for bool type configurations.}
\label{fig:bool_prompt}
\end{center}
\end{figure*}

\section{Prompt Templates in TuneAgent}
To support structured interaction with the Linux kernel configuration space, TuneAgent uses type-aware prompt templates for different Kconfig option types, including Bool, Menu, Choice, and Value. All templates share the same core structure: they specify the \textbf{tuning target}, present a group of candidate kernel configurations, require the agent to reason in a standardized format, and constrain the final answer to a machine-checkable form. This design enables consistent reasoning, automated validation, and reward computation while respecting the semantics of different option types.

\paragraph{Representative Example: Bool Prompt.}
Figure~\ref{fig:bool_prompt} shows the Bool prompt used for kernel tuning exploration. Given a tuning target and a set of Boolean configuration options, the agent determines whether each option is likely to increase performance, decrease performance, or have no clear impact under the target workload. The output is required to follow a structured format, which allows TuneAgent to verify configuration relevance and compute rule-based rewards automatically. Menu, Choice, and Value prompts follow the same template structure, but adapt the final answer space to selecting relevant menus, choosing one option from a mutually exclusive set, or assigning a valid literal value, respectively.

\section{Dataset Construction Details}
\label{app:dataset-details}

Table~\ref{tab:targets} presents a comprehensive and diverse taxonomy of tuning targets and evaluation metrics that underpin the construction of the TuneAgent dataset. 

By systematically spanning a broad spectrum of core kernel subsystems---including CPU scheduling, memory management, file I/O, networking, and process management---the table ensures that the dataset captures both fine-grained kernel mechanisms and end-to-end system behavior across heterogeneous execution contexts. 

The selected tasks collectively cover a wide range of representative performance bottlenecks observed in real-world workloads. Meanwhile, the associated metrics provide rich, quantitative, and interpretable feedback signals that support both LLM reasoning and RL policy optimization. 

Taken together, this broad and multifaceted set of tuning targets and metrics establishes a principled and data-driven foundation for kernel tuning, enabling TuneAgent to learn structured, performance-aware, and deployable configuration strategies that generalize across diverse workloads and operating conditions.

\begin{table*}[htbp]
\centering
\caption{
Overview of kernel subsystems, representative tuning tasks, and evaluation metrics in the TuneAgent dataset. Each subsystem is paired with optimization targets and measurable indicators used to assess and guide RL-based tuning.
}
\label{tab:targets}
\begin{tabular}{|l|p{6cm}|p{6cm}|}
\hline
\textbf{Subsystem} & \textbf{Tuning Task} & \textbf{Key Metrics} \\ \hline

CPU &
Computation efficiency, scheduler behavior, context switching, branch prediction &
CPU score, context-switch cost, scheduler balance, branch efficiency \\ \hline

Memory &
Allocation/release, page management, memory bandwidth, NUMA access, THP &
Memory bandwidth, allocation latency, NUMA latency, THP impact \\ \hline

File I/O &
VFS performance, page cache, metadata operations, block I/O scheduling &
I/O throughput, cache hit rate, metadata latency, disk access time \\ \hline

Pipe &
Pipe buffer configuration, pipe throughput, inter-process transfer latency &
Pipe read/write speed, buffer utilization, transfer delay \\ \hline

Shell &
Process creation/destruction, command execution, shell memory usage &
Command execution time, memory footprint, system load \\ \hline

System Call &
System call dispatch, argument validation, execution latency, concurrency handling &
System call latency, concurrency efficiency, error rate \\ \hline

Thread/Process &
Thread lifecycle, synchronization, mutex contention, IPC, scheduler load balancing &
Thread creation latency, mutex wait time, IPC latency \\ \hline

Network &
TCP buffer tuning, connection tracking, TCP Fast Open, interrupt handling &
Throughput, RTT, retransmission rate, connection delay \\ \hline

Locking &
Lock contention, spinlock efficiency, atomic operations, critical sections &
Lock acquisition time, atomic throughput, critical-section time \\ \hline

Application Scenarios &
Database workloads, network services, system-call-heavy applications &
Query latency, protocol efficiency, interrupt delay, end-to-end throughput \\ \hline

\end{tabular}
\end{table*}

\section{Experiments}
\subsection*{UnixBench Benchmark}
UnixBench is a standard benchmark suite for evaluating the overall performance of Unix-like systems, including Linux. It covers multiple aspects of system behavior, such as integer and floating-point computation, file I/O, process scheduling, pipe communication, system calls, and memory/cache performance. These workloads make UnixBench suitable for assessing how different kernel configurations affect both computation- and system-level efficiency.

In our evaluation, we use UnixBench to measure the impact of TuneAgent-generated kernel configurations across representative kernel-sensitive workloads. The benchmark reports individual scores for each test and aggregates them into an overall normalized score:
\[
\text{UnixBench Score} =
\frac{\text{Total Test Score}}{\text{Reference Score}} \times 100 .
\]
A higher score indicates better system performance. By comparing UnixBench scores under default and TuneAgent-optimized configurations, we quantify the effectiveness of kernel tuning on overall system performance.





\section{Applications and Benchmarks}

\paragraph{Real-world Applications.}
To evaluate the generalization capability of TuneAgent beyond synthetic or micro-benchmark settings, we further test it on three representative real-world applications with distinct workload characteristics:
\begin{itemize}
    \item \textbf{Nginx}: a high-performance web server and reverse proxy, representing network-intensive web-serving workloads.
    \item \textbf{Redis}: an in-memory key-value store, representing memory-intensive and latency-sensitive workloads.
    \item \textbf{PostgreSQL}: a relational database system, representing database workloads involving memory management, I/O operations, and system calls.
\end{itemize}

\paragraph{Benchmarks.}
For each application, we use a standard benchmark that reflects its typical deployment scenario:
\begin{itemize}
    \item \textbf{ApacheBench for Nginx}: measures HTTP request handling capability, including request throughput and response latency.
    \item \textbf{Redis Benchmark for Redis}: measures command-level throughput and latency for common key-value operations.
    \item \textbf{Sysbench for PostgreSQL}: measures database transaction performance under multi-threaded workloads.
\end{itemize}

\paragraph{Evaluation Metrics.}
We report application-specific performance metrics according to the corresponding benchmark. For Nginx, we focus on HTTP request throughput and response latency. For Redis, we measure command throughput and latency. For PostgreSQL, we report transaction throughput and latency. Together, these metrics capture the impact of TuneAgent on network processing, memory access, I/O efficiency, scheduling behavior, and system-call overhead in realistic deployment environments.

\section{Limitations and Future Work}

TuneAgent focuses on Linux kernel tuning because Linux is open source and provides a mature, accessible, and well-documented configuration framework. This makes it suitable for systematic optimization and reproducible evaluation, but also limits our current study to Linux-based systems. Extending TuneAgent to closed-source OS remains difficult because their kernel source code, configuration dependencies, and tuning interfaces are often inaccessible. Even within Linux, configuration spaces may vary across kernel versions, distributions, and hardware platforms, which can affect the transferability of learned tuning policies.

Another limitation is the use of a proxy evaluator for performance-aware rewards during training. Since rebuilding, deploying, and benchmarking every candidate configuration is expensive, the proxy reward provides efficient training-time feedback. However, its estimates may deviate from real benchmark outcomes under unseen workloads or hardware. Therefore, TuneAgent uses the proxy reward only for training-time guidance, while all final performance results are obtained through real system benchmarking. 

Future work can further extend TuneAgent in several directions. First, TuneAgent can be adapted to broader hardware architectures, such as ARM and RISC-V, and evaluated across different kernel versions and distributions. Second, future systems could support real-time adaptation to dynamic workloads, which is important for cloud and edge environments. Third, incorporating low-level diagnostic signals, such as hardware counters, tracing information, and system logs, could provide finer-grained feedback for more accurate tuning. Finally, future work may explore multi-objective optimization, human-in-the-loop refinement, and large-scale deployment across clusters or heterogeneous systems.